\documentclass[12pt]{iopart}

\usepackage{graphicx}
\usepackage[export]{adjustbox}
\expandafter\let\csname equation*\endcsname\relax
\expandafter\let\csname endequation*\endcsname\relax
\usepackage{amsfonts}
\usepackage{amsmath}
\usepackage{algorithmic}
\usepackage{array}
\usepackage[caption=false,font=normalsize,labelfont=sf,textfont=sf]{subfig}
\usepackage{textcomp}
\usepackage{stfloats}
\usepackage{url}
\usepackage{verbatim}
\usepackage{balance}
\usepackage{color}
\usepackage{float}
\usepackage{hyperref}
\makeatletter
\renewcommand{\@makecaption}[2]{%
  \vskip\abovecaptionskip
  \sbox\@tempboxa{\footnotesize\textbf{#1:} \footnotesize{#2}}%
  \ifdim \wd\@tempboxa >\hsize
    \footnotesize\textbf{#1:} \footnotesize{#2}\par
  \else
    \global \@minipagefalse
    \hb@xt@\hsize{\hfil\box\@tempboxa\hfil}%
  \fi
  \vskip\belowcaptionskip}
\makeatother

\begin{document}

\title[Understanding the Functional Roles of Modelling Components in SNN]{Understanding the Functional Roles of Modelling Components in Spiking Neural Networks}

\author{Huifeng Yin\textsuperscript{1}, Hanle Zheng\textsuperscript{1}, Jiayi Mao\textsuperscript{1}, Siyuan Ding\textsuperscript{2}, Xing Liu\textsuperscript{3}, Mingkun Xu\textsuperscript{4}, Yifan Hu\textsuperscript{1}, Jing Pei\textsuperscript{1}, Lei Deng\textsuperscript{1,*}}

\address{\textsuperscript{1} Center for Brain Inspired Computing Research (CBICR), Department of Precision Instrument, Tsinghua University, Beijing, China.}
\address{\textsuperscript{2} Weiyang College, Tsinghua University, Beijing, China.}
\address{\textsuperscript{3} College of Electronic Information and Automation, Tianjin University of Science and Technology, Tianjin, China.}
\address{\textsuperscript{4} Guangdong Institute of Intelligence Science and Technology, Zhuhai, China.}
\address{\textsuperscript{*} Author to whom any correspondence should be addressed.}
\ead{leideng@mail.tsinghua.edu.cn}
\vspace{10pt}
\begin{indented}
\item[]December 2023
\end{indented}


\begin{abstract}

Spiking neural networks (SNNs), inspired by the neural circuits of the brain, are promising in achieving high computational efficiency with biological fidelity. Nevertheless, it is quite difficult to optimize SNNs because the functional roles of their modelling components remain unclear. By designing and evaluating several variants of the classic model, we systematically investigate the functional roles of key modelling components, leakage, reset, and recurrence, in leaky integrate-and-fire (LIF) based SNNs. Through extensive experiments, we demonstrate how these components influence the accuracy, generalization, and robustness of SNNs. Specifically, we find that the leakage plays a crucial role in balancing memory retention and robustness, the reset mechanism is essential for uninterrupted temporal processing and computational efficiency, and the recurrence enriches the capability to model complex dynamics at a cost of robustness degradation. With these interesting observations, we provide optimization suggestions for enhancing the performance of SNNs in different scenarios. This work deepens the understanding of how SNNs work, which offers valuable guidance for the development of more effective and robust neuromorphic models.

\end{abstract}

\section{Introduction}


The human brain's extraordinary computing capabilities have intrigued researchers for centuries. While deep artificial neural networks (ANNs) have recently made strides in emulating brain functions \cite{hassabis2017neuroscience, lecun2015deep, he2016deep, tai2015improved, vaswani2017attention}, they present limitations in capturing the brain's rich temporal dynamics and high energy efficiency \cite{deng2020rethinking, marblestone2016toward, tavanaei2019deep}. These limitations are alleviated by another family of neural networks, termed as spiking neural networks (SNNs) \cite{maass1997networks, zhou2022exploring}, which can reach more closely the brain's capability in processing information through spatio-temporal encoding while being energy efficient \cite{han2020deep}. SNNs offer a promising avenue for achieving better biological fidelity and computational efficiency in neural modelling. 

Other key advantages of SNNs lie in their robustness and generalization. SNNs present robust performance when generalizing across diverse data types and conditions. For instance, when handling data collected by dynamic vision sensors (DVS) \cite{amir2017low} with varying temporal resolutions, SNNs have demonstrated superior recognition accuracy compared to recurrent neural networks (RNNs) \cite{he2020comparing}. Moreover, SNNs exhibit remarkable robustness in resisting adversarial attacks, as Liang et al. demonstrated that attacking SNN models needs larger perturbations than attacking ANN models \cite{liang2021exploring}. These advantages are intrinsically related to the complex temporal dynamics and firing mechanism of SNNs.

However, the optimization of SNNs is not easy because the functional roles of their modelling components are quite unclear \cite{liang2021h2learn, bellec2018long}. Taking the most widely used SNN model, the leaky integrate-and-fire (LIF) model \cite{izhikevich2003simple,gerstner2014neuronal}, as an example, it includes several modelling components such as the membrane potential dynamics, the leakage of the membrane potential, and the spike generation mechanism. Recent studies have advanced our understanding of the roles of individual modelling components in LIF-based SNN models, yet the functional roles of all these components and their impacts on the model performance remain under-explored. For example, Bouanane et al. found that the leakage of spiking neuron models in feedforward networks does not necessarily lead to improved performance, even in processing temporally complex tasks \cite{bouanane2022impact}. Similarly, Chowdhury et al. highlighted the trade-off involved in incorporating the leaky behavior in neuron models, particularly focusing on the balance between the computational efficiency and robustness against noisy inputs \cite{chowdhury2021towards}. Yao et al. concluded that the hard reset mechanism reducing the current membrane potential to an empirical value, e.g., 0, restarts the potential trace and provides a stable neuronal dynamics \cite{yao2022glif}. Ponghiran and Roy revealed the limitations of inherent recurrence in conventional SNNs for sequential learning, and then modified it for enhanced long-sequence learning \cite{ponghiran2022spiking}. These studies have partially explored the functionalities of specific LIF components but lack a systematic identification and in-depth analysis of all components, which is insufficient for understanding and even optimizing LIF-based SNN models.

To bridge this gap, this work aims at understanding the functional roles of modelling components in LIF-based SNNs systematically and inspiring the optimization strategies with the findings. The modelling components in this work include the leakage component, the reset component, and the recurrence component in LIF-based SNNs. By constructing several LIF variants with different component combinations and examining their performance on real-world benchmarks, we demonstrate their contributions to the SNN performance, such as accuracy, generalization, and robustness. Our first objective is to deepen our understanding of the functional roles of different modelling components in LIF-based SNNs. Furthermore, we discuss the model optimization suggestions for different tasks based on the observations from above comprehensive experiments. Notice that we do not remove the firing mechanism to maintain the models as SNNs in which neurons communicate with each other using spikes.

The remainder of this paper is organized as follows. Section 2 provides a brief overview of SNNs, particularly focusing on the LIF-based SNN model and its key modelling components. Section 3 outlines the approach, detailing the design of variant models to identify the functional roles of these components. In Section 4, we present and analyze the results in extensive experiments, exploring the impacts of different modelling components on performance, generalization, and robustness. Section 5 discusses the model optimization suggestions for SNNs based on the observations gained from the experiments. Finally, Section 6 concludes this work.

\section{Spiking Neural Networks: A Brief Overview}

\subsection{LIF-based SNN Model}

\begin{figure}[ht]
\centering
\includegraphics[width=\linewidth]{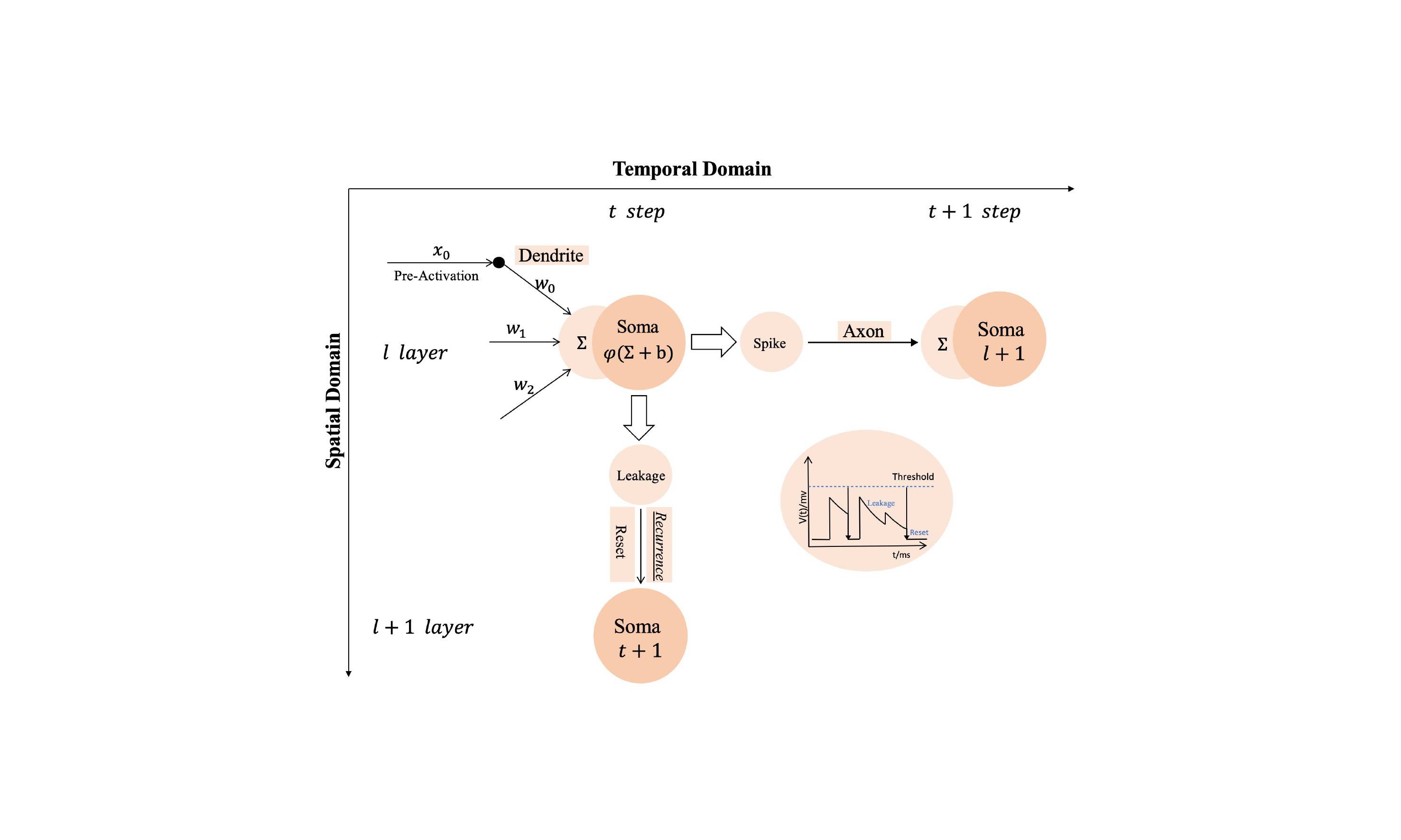}
\caption{Illustration of the LIF-based spiking neuron model: (a) spike integration, transformation, and generation in a spiking neuron; (b) network structure of SNNs highlighting temporal and spatial dimensions; (c) leakage component reflecting membrane potential decaying; (d) reset component reflecting the membrane potential reset after each spike; (e) recurrence reflecting cross-neuron influence in the temporal dimension.}
\label{fig:lif}
\end{figure}

The LIF-based spiking neuron model is a simplistic representation of a biological neuron, which is widely used as a classic format in SNN modelling. LIF-based SNN models aim to emulate the behaviors of the brain with accurate functional emulation and high computational efficiency by leveraging the temporal dynamics of a spiking neuron. The LIF neuron model is illustrated in Figure~\ref{fig:lif}.

The LIF neuron model is governed by two main parts: the membrane potential dynamics and the spike generation mechanism. The membrane potential $u(t)$ can be described by the following differential equation:
\begin{equation}
\tau \frac{d u}{d t} = -[u(t) - u_{\text{rest}}] + R I(t),
\end{equation}
where $\tau$ is the membrane time constant, $u_{\text{rest}}$ is the resting membrane potential, $R$ is the membrane resistance, and $I(t)$ represents the input current. When the membrane potential exceeds a certain threshold value $u_{th}$, the neuron would fire a spike, and the membrane potential is reset to $u_{\text{reset}}$. This process can be formulated as
\begin{equation}
\begin{cases}o(t)=1~\&~u(t)=u_{reset}, & \text { if } u(t) \geq u_{t h} \\ o(t)=0, & \text { if } u(t)<u_{th}\end{cases}
\end{equation}
where $o(t)$ represents the output spike at the $t$-th time step.

The continuous LIF model can be discretized using the Euler method, yielding the following iterative LIF model like
\begin{equation}
\begin{cases}
u_i^{t, n} & =k_\tau u_i^{t-1, n}\left(1-o_i^{t-1, n}\right)+\sum_j w_{i j}^n o_j^{t, n-1}, \\
o_i^{t, n} & =H\left(u_i^{t, n}-u_{t h}\right),
\end{cases} 
\end{equation}
where $u_i^{t, n}$ is updated at each time step in the $n$-th layer based on its previous state and the current input spikes. The synaptic weight connecting neurons $i$ and $j$ is denoted by $w_{ij}^n$, and $o_i^{t, n}$ represents the output spike of neuron $i$ at the $t$-th time step. The model also incorporates a leakage coefficient, $k_\tau$, to simulate the gradual decaying effect of the membrane potential over time, and a threshold potential, $u_{th}$, to control whether the neuron fires a spike. $H(\cdot)$ denotes the classic Heaviside function. 

\subsection{LIF Modelling Components}

\subsubsection{Leakage Component.}
$k_\tau$ represents the leakage coefficient ranging within $[0,~1]$, which determines how fast the membrane potential $u$ decays over time. This component is responsible for the ``leaky'' functionality of the LIF neuron model, as it simulates the passive decay effect of the membrane potential.

\subsubsection{Reset Component.}
The output spike $o$ reflects the firing mechanism of the membrane potential and $(1-o)$ is responsible for resetting the membrane potential once the neuron fires. When the neuron fires (i.e., $o = 1$), the reset mechanism resets the membrane potential by multiplying it by zero (i.e., $(1-o) = 0$) in updating the membrane potential at the next time step. Otherwise, the membrane potential will remain unchanged. This component returns the membrane potential to a reset state after each spike event.

\subsubsection{Recurrence Component.}
This component is not presented in the equations above but can be added as recurrent connections at the network level. When included, the membrane potential dynamics can be rewritten as
\begin{equation}
u_i^{t, n}=k_\tau u_i^{t-1, n}\left(1-o_i^{t-1, n}\right)+\sum_j v_{i j}^n o_j^{t-1, n}+\sum_j w_{i j}^n o_j^{t, n-1}
\end{equation}
where $v_{i j}^n$ represents the strength of the recurrent connection (synaptic weight) from neuron $j$ to neuron $i$ in the $n$-th layer. $o_j^{t-1, n}$ represents the spike state of neuron $j$ at the $(t-1)$-th time step. The recurrence component essentially provides feedback signals from the previous time step, allowing neurons to take the historic spike states of other neurons in the same layer into account when updating their membrane potentials.

\section{Approach} 

\begin{figure}[ht]
\centering
\includegraphics[width=\linewidth]{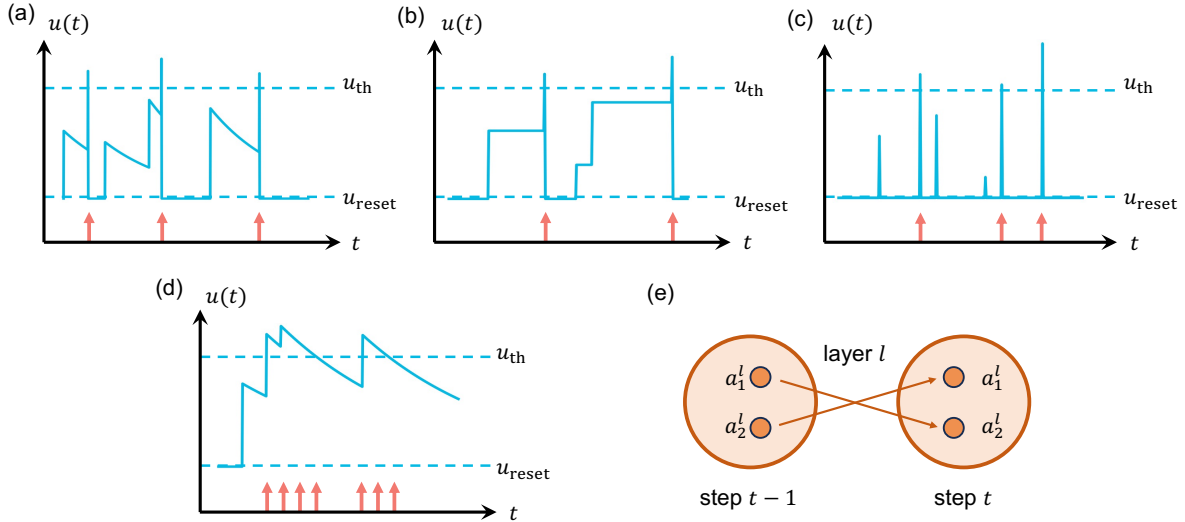}
\caption{Variant models of LIF-based SNNs: (a) vanilla LIF; (b) without leakage; (c) complete leakage; (d) without reset; (e) with recurrence.}
\label{fig:components}
\end{figure}

In this section, we design several variant models with different component combinations as shown in Figure~\ref{fig:components}, allowing us to investigate the functional roles of different modelling components in LIF-based SNNs. While these variants may be biologically implausible, they provide valuable insights by contrasting the effects of specific modifications. This approach follows a common ablation study method, which systematically examines the impact of including or excluding each component in the baseline model. The primary intention of this work is not to ensure the biological plausibility of these variants but to understand the functional roles of the modelling components of SNNs.

\subsection{Different Leakage Coefficients}

The leakage component determines the decaying rate of the neuron's membrane potential. This is an essential property of LIF neurons, as it can prevent the membrane potential from accumulating indefinitely. We design three variants as follows for considering this component.

\textbf{Normal leakage}. In this case, the membrane potential decays with a normal rate:
\begin{equation}
u_i^{t, n} =k_\tau u_i^{t-1, n}\left(1-o_i^{t-1, n}\right)+\sum_j w_{i j}^n o_j^{t, n-1},
\end{equation}
which represents a balance between the cases without and complete leakage. This setting follows the vanilla LIF model with better bio-plausibility. Here we usually set $k_\tau=0.3$ for a typical value.

\textbf{Without leakage}. In this variant, the leakage coefficient is set to one:
\begin{equation}
u_i^{t, n} = 1\cdot u_i^{t-1, n}\left(1-o_i^{t-1, n}\right)+\sum_j w_{i j}^n o_j^{t, n-1},
\end{equation}
which means that the membrane potential does not decay at all. This results in full integration of current input signals and the previous membrane potential, allowing the neuron to accumulate inputs indefinitely.

\textbf{Complete leakage}. In this variant, the leakage coefficient is set to zero:
\begin{equation}
u_i^{t, n} = 0\cdot u_i^{t-1, n}\left(1-o_i^{t-1, n}\right)+\sum_j w_{i j}^n o_j^{t, n-1}=\sum_j w_{i j}^n o_j^{t, n-1},
\end{equation}
which means that the membrane potential decays completely at each single time step and is only determined by the current inputs.

\subsection{Different Reset Modes}

The reset component is responsible for resetting the membrane potential of the neuron every time it fires, which prevents the neuron from firing continuously with a large potential value. We design two variants as follows for considering this component.

\textbf{Normal reset}. In this case, the reset term is included in the equation:
\begin{equation}
u_i^{t, n} =k_\tau u_i^{t-1, n}\left(1-o_i^{t-1, n}\right)+\sum_j w_{i j}^n o_j^{t, n-1}.
\end{equation}
This reset term ensures that the neuron's membrane potential will be reset to a lower state if the neuron fired in the last time step, i.e., $o_i^{t-1, n} = 1$.

\textbf{Without reset}. In this variant, the reset term is removed from the equation:
\begin{equation}
u_i^{t, n} =k_\tau u_i^{t-1, n}+\sum_j w_{i j}^n o_j^{t, n-1}.
\end{equation}
Consequently, the membrane potential will not be reset after it fires, allowing the neuron to continuously fire once its membrane potential reaches the threshold. This can lead to different dynamical properties and firing rates of the network.\\

\subsection{Different Recurrence Patterns}

The recurrence component introduces recurrent connections within the network, allowing neurons to influence each other's states. We design two variants as follows for considering this component.

\textbf{Without recurrence}. In this case, there are no recurrent connections between neurons:
\begin{equation}
u_i^{t, n} =k_\tau u_i^{t-1, n}\left(1-o_i^{t-1, n}\right)+\sum_j w_{i j}^n o_j^{t, n-1}.
\end{equation}
The network is fully feedforward, and the output of one neuron does not influence the states of other neurons within the same layer at different time steps.

\textbf{With recurrence}. In this variant, recurrent connections between neurons are introduced into the network, as specified by the $\sum_{j} \mathrm{v}_{i j}^n o_{j}^{t-1, n}$ term: 
\begin{equation}
u_i^{t, n} =k_\tau u_i^{t-1, n}\left(1-o_i^{t-1, n}\right)+\sum_j w_{i j}^n o_j^{t, n-1}+\sum_{j} \mathrm{v}_{i j}^n o_{j}^{t-1, n}.
\end{equation}
This allows neurons to influence each other's states in a feedback loop, leading to more complex dynamics and better learning of temporal features.

\subsection{Implementation on Neuromoprhic Hardware}

Our modifications on the LIF model do not inherently complicate their implementation on neuromorphic hardware. For instance, the Tianjic chip \cite{pei2019towards, deng2020tianjic}, which is elaborated for both ANNs and SNNs with a hybrid architecture, can accommodate the models used in this work. By configuring the leakage coefficient, the reset mode and the connection topology between neurons, it is easy to deploy the proposed variant models.

\section{Experiments}

This section offers an in-depth exploration of the functional roles of modelling components in LIF-based SNNs through a series of evaluation experiments, aiming at understanding how leakage, reset, and recurrence mechanisms impact the model performance. We initially focus on investigating the functional roles of three core components on different types of benchmarks. Our findings indicate that the impacts of these components can vary significantly on different datasets. Then, we further explore the generalization and robustness of SNNs. The generalization experiments focus on the adaptability of SNNs to different types of unseen data, with a particular emphasis on the neuromorphic datasets collected by DVS. In addition, the robustness assessment concentrates on the SNNs' ability to resist adversarial attacks, serving as a complementary confirmation of the generalization capability. These comprehensive experiments help deepen our understanding of how SNNs work, paving the way for future development of neuromorphic models.

\subsection{Accuracy Analysis on Different Benchmarks}

\subsubsection{Selection of Benchmarks.}

In order to evaluate the performance of SNNs extensively, particularly focusing on the model variants of LIF-based SNNs, we select a range of benchmarks. These benchmarks are selected for their diversity and complexity, which enable a comprehensive assessment of different modelling components of SNNs and their functional roles.

\textbf{Delayed Spiking XOR Problem.} The delayed spiking XOR problem is customized to test the long-term memory capabilities of different neural network models \cite{zheng2024temporal}. This problem is structured in three stages: initially, an input spike pattern with a varying firing rate is injected into the network; then, it is followed by a prolonged delay period filled with noisy spikes; finally, the network receives another spike pattern and the network is expected to output the result of an XOR operation between the initial and final input spike patterns. The XOR operation is a concept in digital circuits, whose input and output signals only have binary states, one or zero. In the delayed spiking XOR problem, we correspond the high-firing-rate and low-firing-rate spike patterns to one and zero, respectively. This benchmark is crucial for understanding how different models memorize long-term information, particularly in scenarios where there is a long-term period filled with irrelevant data between two critical pieces of information. 

\textbf{Temporal Datasets.} For temporal datasets, we select two speech signal datasets, SHD and SSC \cite{cramer2020heidelberg}, listed in Table~\ref{tab:temporal datasets}, as they provide rich temporal information that can test the capability of SNNs in processing temporal dependencies.

\begin{table}[ht]
\centering
\footnotesize
\caption{Information of temporal datasets.}
\begin{tabular}{lll}
\br
Attribute & SHD & SSC \\
\mr
Recordings & 10,420 & 105,829 \\
Classes & 10 & 35 \\
Type & Audio, spiking data & Audio, spiking data \\
Source & Microphone array & Microphone array \\
Format & Spike times & Spike times \\
\br
\end{tabular}
\label{tab:temporal datasets}
\end{table}

\textbf{Spatial Datasets.} For the assessment of SNNs in handling spatial information, we select datasets listed in Table~\ref{tab:spatial datasets} whose data are distributed spatially in nature. This includes the MNIST dataset \cite{lecun1998gradient}, famous for its collection of grayscale images of handwritten digits, and the CIFAR10 dataset \cite{krizhevsky2009learning}, which presents a more complex set of color images depicting various natural scenes and objects. These spatial datasets are crucial to understand how SNNs can perform in recognizing spatial features.

\begin{table}[ht]
\centering
\footnotesize
\caption{Information of spatial datasets.}
\begin{tabular}{lll}
\br
Attribute & MNIST & CIFAR10 \\
\mr
Images & 70,000 & 60,000 \\
Classes & 10 & 10 \\
Size & 28x28 pixels & 32x32 pixels \\
Source & Handwritten digits & Natural images \\
Format & Grayscale & RGB \\
\br
\end{tabular}
\label{tab:spatial datasets}
\end{table}

\textbf{Spatio-Temporal Datasets.} To bridge the gap between purely temporal and spatial datasets, we additionally select datasets collected by DVS cameras, including Neuromorphic-MNIST (N-MNIST) \cite{orchard2015converting} and DVS128 Gesture \cite{amir2017low}, listed in Table~\ref{tab:spatio-temporal datasets}. DVS can capture visual information in a dynamic and event-driven manner, offering a blend of information in both spatial and temporal dimensions. The N-MNIST dataset, a neuromorphic adaptation of the spatial MNIST dataset, presents handwritten digits in a sequential temporal format. The DVS128 Gesture dataset, conversely, comprises recordings of various hand gestures, showcasing complex spatio-temporal patterns. These datasets provide a way to evaluate the capability of SNNs in handling spatio-temporal features simultaneously, making them ideal for understanding the functional roles of their modelling components. For the N-MNIST dataset, we use a fixed time interval to integrate events as frames for post-processing with a limited number of time steps. However, for the DVS128 Gesture dataset, due to the variation in sample durations, we adopt a fixed-frame compression method to ensure that all samples are of uniform length \cite{fang2021incorporating, doi:10.1126/sciadv.adi1480}.

\begin{table}[ht]
\centering
\footnotesize
\caption{Information of spatio-temporal datasets.}
\begin{tabular}{lll}
\br
Attribute & N-MNIST & DVS128 Gesture \\
\mr
Samples & 70,000 & 1,342 \\
Classes & 10 & 11 \\
Size & 34x34 pixels & 128x128 pixels \\
Source & Neuromorphic handwritten digits & Dynamic hand gestures \\
Format & DVS format & DVS format \\
\br
\end{tabular}
\label{tab:spatio-temporal datasets}
\end{table}

\subsubsection{Experimental Setup}

We implement the LIF neuron model with different configurations for each of the three modelling components: leakage, reset, and recurrence. The performance of all variant models is evaluated on selected benchmarks. We employ similar network structures and hyper-parameter configurations for different variant models to provide a fair comparison. All networks are trained with backpropagation through time (BPTT) \cite{wu2018spatio, wu2019direct, lian2023learnable, neftci2019surrogate, wang2023adaptive, zenke2021remarkable}, in which surrogate gradients are used to solve the nondifferentiability of spike activities. The detailed network architectures for each dataset are provided in Table \ref{tab:network structures}. Note that the complexity of CIFAR10 makes it challenging for Multi-Layer Perceptrons (MLPs). To overcome this issue, we employ SNNs with convolutional layers. Therefore, experiments with recurrence are omitted for CIFAR10, given the inconvenience in adding recurrent connections onto convolutional architectures\cite{lecun1995convolutional, he2016deep}.

\begin{table}[ht]
\centering
\caption{Network structures for different models and datasets.}
\resizebox{\textwidth}{!}{
\begin{tabular}{@{}ll}
\br
Dataset & Network structures \\
\mr
SHD & Input(700)--$\textnormal{FC}_{LIF}$(64)-$\textnormal{FC}_{LIF}$(20)-Output  \\
SSC & Input(700)--$\textnormal{FC}_{LIF}$(200)-$\textnormal{FC}_{LIF}$(35)-Output \\
N-MNIST & Input(2312)--$\textnormal{FC}_{LIF}$(512)-$\textnormal{FC}_{LIF}$(10)-Output  \\
DVS128 Gesture & Input(32768)--Downsampling(2048)--$\textnormal{FC}_{LIF}$(512)-$\textnormal{FC}_{LIF}$(512)-$\textnormal{FC}_{LIF}$(11)-Output  \\
MNIST & Input(784)--$\textnormal{FC}_{LIF}$(512)-$\textnormal{FC}_{LIF}$(10)-Output  \\
CIFAR10 & Input(3072)--$\textnormal{ResNet18}_{LIF}$-Output  \\
\br
\end{tabular}}
\label{tab:network structures}
\end{table}

We adopt a uniform set of hyper-parameters for different models, ensuring a fair comparison between variant models. These hyper-parameters include the number of epochs, batch size, learning rate, and SNN-specific parameters like the firing threshold and the gradient width during backpropagation. The Adam optimizer is employed for all models, with a learning rate scheduler to adjust the learning rate during training. The detailed hyper-parameter settings can be found in Table~\ref{tab:hyperparameter}, enabling a consistent and reproducible experimental setup.

\begin{table}[ht]
\footnotesize
\centering
\caption{Hyper-parameter settings for different models and datasets.}
\resizebox{\textwidth}{!}{%
\begin{tabular}{lccccccccc}
\hline
& \multicolumn{3}{c}{Temporal} && \multicolumn{2}{c}{Spatio-Temporal} && \multicolumn{2}{c}{Spatial} \\
\cline{2-4} \cline{6-7} \cline{9-10}
Dataset & XOR & SHD & SSC && N-MNIST & DVS128 Gesture && MNIST & CIFAR10 \\
\hline
\#Epochs & 150 & 100 & 100 && 100 & 100 && 100 & 100 \\
Batch Size & 500 & 100 & 100 && 512 & 512 && 512 & 512 \\
Learning Rate & 1e-2 & 1e-2 & 1e-2 && 1e-4 & 0.1 && 1e-4 & 1e-4 \\
\hline
$u_{th}$ & 0.5 & 0.5 & 0.5 && 0.3 & 0.3 && 0.3 & 0.3 \\
Surrogate Gradient Width & 0.5 & 0.5 & 0.5 && 0.25 & 0.25 && 0.25 & 0.25 \\
\hline
Optimizer & Adam & Adam & Adam && Adam & Adam && Adam & Adam \\
Scheduler & StepLR (50, 0.1) & StepLR (20, 0.5) & StepLR (25, 0.1) && StepLR (25, 0.1) & StepLR (25, 0.1) && StepLR (25, 0.1) & StepLR (25, 0.1) \\
\hline
\end{tabular}}
\label{tab:hyperparameter}
\end{table}

Specifically, for temporal datasets where temporal dependencies are quite hard to learn, we adopt a learnable leakage in the baseline for better accuracy. In contrast, for spatio-temporal and spatial datasets, we maintain a fixed leakage across all experiments. In all experiments, each variant model alters only one component at a time, ensuring that the observed effects can be attributed solely to the change of the specific component.

\subsubsection{Results and Analyses.}

We conduct comprehensive experiments to compare the performance of different variants on different datasets. It should be noted that we do not intend to design a better model, but to study the functional roles of different components. As reflected in Table \ref{tab:acc}, we observe that the influence of the three modelling components of SNNs, leakage, reset, and recurrence, varies significantly on different types of benchmarks. Notably, the impact is the most significant on temporal benchmarks, followed by spatio-temporal datasets and spatial datasets. This trend aligns with reasonable expectations considering the nature of these benchmarks. Temporal ones inherently involve dynamic dependencies over time, thereby highlighting the effects of these components more significantly in adapting the membrane potential dynamics. Spatio-temporal datasets with hybridization of temporal dynamics and spatial features, exhibit a moderate level of influence. In contrast, spatial datasets, which primarily focus on static spatial features, demonstrate the least sensitivity to these variations in component combinations. Therefore, we recommend using the results on temporal and spatio-temporal datasets for in-depth analyses to avoid misleading in this part.

\begin{table*}[ht]
\centering
\caption{Accuracy comparison between variant models on different benchmarks.}
\resizebox{\textwidth}{!}{
\begin{tabular}{lccccccccc}
\br
& \multicolumn{3}{c}{Temporal} && \multicolumn{2}{c}{Spatio-Temporal} && \multicolumn{2}{c}{Spatial} \\
\cline{2-4} \cline{6-7} \cline{9-10}
Dataset & XOR & SHD & SSC && N-MNIST & DVS128 Gesture && MNIST & CIFAR10 \\
\mr
Baseline  & 77.50\%  & 78.58\% & 60.70\% && 94.75\% & 94.09\% && 94.85\% & 93.09\% \\
Without leakage    & 50.50\%  & 13.69\% & 3.22\% && 95.18\% & 89.23\% && 94.38\% & 92.97\% \\
Complete leakage       & 50.30\%  & 67.67\%  & 44.45\%  && 95.03\% & 93.75\% && 95.22\% & 92.25\% \\
Without reset & 96.10\% & 78.89\% & 60.70\% && 96.18\% & 93.75\% && 94.75\% & 93.29\% \\
With recurrence & 98.90\% & 80.43\% & 66.53\% && 96.21\% & 93.89\% && 96.98\% & -- \\
\br
\end{tabular}}
\label{tab:acc}
\end{table*}

\textbf{The role of the leakage.} The leakage coefficient directly determines the decaying rate of the membrane potential over time. In processing long sequences, the decaying rate of the membrane potential is crucial for the model capability in learning long-term temporal dependencies. A appropriate leakage rate generally allows the model to handle temporal information more effectively.

Specifically, a model configured without leakage (i.e., $k_\tau=1$) retains full membrane potentials from the previous time step, which hinders the model from responding to current inputs swiftly. This leads to poor performance especially on temporal benchmarks, with classification accuracy barely above random guessing. Conversely, complete leakage (i.e., $k_\tau=0$) makes the model fail to retain any information from previous time steps, thus also degrading the model accuracy. The optimal performance is achieved with a proper leakage rate setting, where the model balances better in retaining and forgetting historical information.

The leakage coefficient is intrinsically linked to the signal frequency that the model can respond well to. As demonstrated in \cite{zheng2024temporal}, a properly high leakage of the LIF model allows the neuronal membrane potential to decay its historic information more swiftly, thereby enabling a quicker adaption to high-frequency inputs which usually change fast. However, a lower leakage means a slow decay of historic information, thus retaining a long-term memory but impairing its sensitive response to high-frequency signals.

\begin{table*}[ht]
\centering
\caption{Accuracy comparison between variant models on different benchmarks.}
\resizebox{\textwidth}{!}{
\begin{tabular}{lcccccccc}
\hline
& \multicolumn{2}{c}{Temporal} && \multicolumn{2}{c}{Spatio-Temporal} && \multicolumn{2}{c}{Spatial} \\
\cline{2-3} \cline{5-6} \cline{8-9}
Dataset & SHD & SSC && N-MNIST & DVS128 Gesture && MNIST & CIFAR10 \\
\hline
$k_\tau = 0$     & 67.67\% & 44.45\% && 95.03\% & 93.75\% && 95.22\% & 92.25\% \\
$k_\tau = 0.1$   & 69.04\% & 45.30\% && 95.04\% & 93.05\% && 95.12\% & 92.67\% \\
$k_\tau = 0.2$   & 67.89\% & 46.53\% && 94.84\% & 93.75\% && 94.77\% & 92.28\% \\
$k_\tau = 0.3$   & 68.29\% & 47.98\% && 94.75\% & 94.09\% && 94.85\% & 93.09\% \\
$k_\tau = 0.4$   & 69.17\% & 49.88\% && 94.69\% & 94.44\% && 95.98\% & 93.98\% \\
$k_\tau = 0.5$   & 72.35\% & 51.69\% && 94.60\% & 93.75\% && 95.62\% & 93.23\% \\
$k_\tau = 0.6$   & 67.76\% & 52.62\% && 94.69\% & 93.40\% && 96.02\% & 93.64\% \\
$k_\tau = 0.7$   & 67.89\% & 54.50\% && 94.84\% & 92.01\% && 95.17\% & 93.21\% \\
$k_\tau = 0.8$   & 63.91\% & 56.52\% && 95.19\% & 90.62\% && 94.85\% & 93.22\% \\
$k_\tau = 0.9$   & 58.61\% & 58.16\% && 95.34\% & 89.93\% && 96.48\% & 92.76\% \\
$k_\tau = 1.0$   & 13.69\% & 3.22\%  && 95.18\% & 89.23\% && 94.38\% & 92.97\% \\
\hline
\end{tabular}}
\label{tab:leakage0.1}
\end{table*}

We conduct a detailed experiment to explore the effect of varying the leakage value from 0 to 1 with an interval of 0.1. The results are presented in Figure \ref{tab:leakage0.1}. First, it is consistent that the impact is more significant on temporal datasets with richer temporal dependencies. Second, a proper leakage configuration, not too high without long-term memory or too low without sensitive response, is helpful for capturing multi-timescale temporal features on temporal datasets.

\textbf{The role of the reset mode.} The reset mechanism of the LIF neuron model serves as another critical role in learning temporal dynamics. When the membrane potential surpasses a certain threshold, the neuron fires a spike, and the reset mechanism subsequently reinitializes the membrane potential to a lower value. While this mechanism aligns with biological plausibility and aims to prevent unbounded potential accumulation, it inadvertently disrupts the temporal continuity that matters in performing certain tasks. 

Specifically, in the delayed spiking XOR problem, the reset mechanism in the baseline model shows a significant degradation in accuracy. The reset process erases the membrane potential completely, cleaning all historic information, which makes it failed to memorize long-term dependencies in this task. This effect is weakened on spatio-temporal and purely spatial benchmarks due to the poorer temporal information. Notice that the impact of the reset mechanism is smaller than that of the leakage component on SHD and SSD datasets with higher complexity, therefore the observed differences are not significant.

\begin{figure}[ht]
\centering
\includegraphics[width=\linewidth]{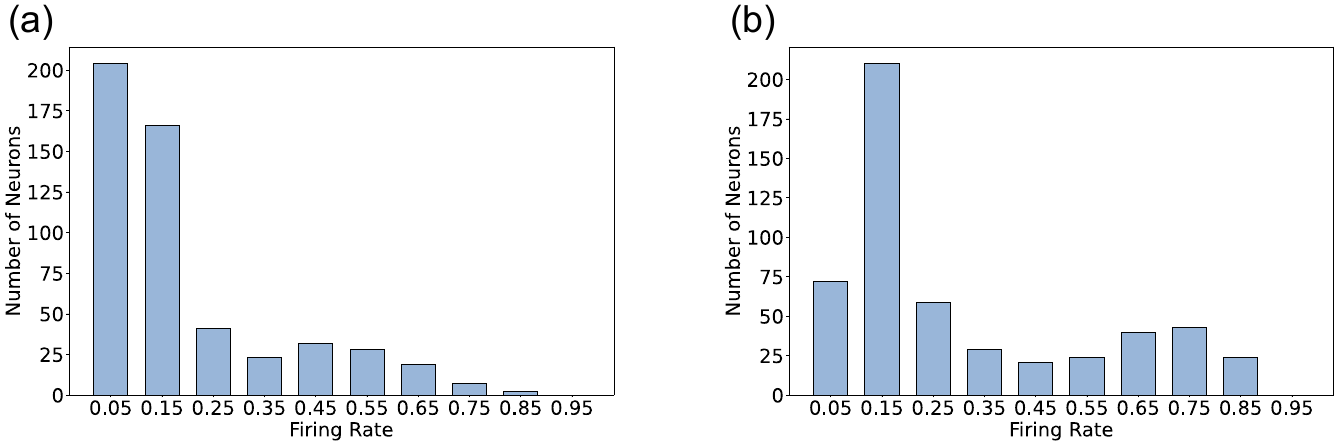}
\caption{Average spike rate distribution of variant models on N-MNIST: (a) normal reset; (b) without reset. }
\label{fig:firing rate}
\end{figure}

In addition, the reset mechanism greatly impacts the firing rate of neurons. As illustrated in Figure~\ref{fig:firing rate}, the reset can reduce the firing rate by returning the membrane potential to a lower value every time a spike fires. This action leads to sparser spike activities, which is advantageous for higher computational efficiency~\cite{deng2020model, deng2021comprehensive}. Such a feature is particularly beneficial in edge computing devices, where computational resources and power supply are limited.

\textbf{The role of the recurrence pattern.} The incorporation of the recurrence component in SNNs endows these models with the capability to exchange information between different neurons in the same layer across time steps. This cross-neuron information fusion significantly enhances the model's capability in learning, capturing, and integrating temporal features. In temporal computing tasks, the presence of recurrence generally results in superior accuracy. Similarly, the improvement would decrease when the benchmarks have fewer temporal dependencies such as on spatial benchmarks.

\subsection{Generalization Analyses on Spatio-Temporal datasets}

Generalization is pivotal in understanding the model's capacity to adapt to unseen data and really matters in real-world applications. Recent studies investigating the generalization capability of SNNs have unveiled noteworthy results, especially in comparison to RNNs \cite{he2020comparing}. In this section, we present a series of experiments on the spatio-temporal datasets collected by DVS, to evaluate the generalization of variant models and analyze the underlying mechanisms. We delve into this phenomenon from two perspectives: the relationship between the flatness of the loss landscapes and the generalization in machine learning theory; the impact of different modelling components of SNNs on error accumulation and gradient backpropagation.

\subsubsection{Experimental Setup}

DVS cameras represent an emerging imaging technology that capture pixel-level changes in luminance, resulting in asynchronous, event-driven, sparse, and temporal event streams. One of the prominent datasets collected by DVS is the N-MNIST dataset, a neuromorphic version of the classic MNIST dataset. It is generated by moving the MNIST images in front of a DVS camera that records the spike outputs, thus converting static images into temporal event streams and offering richer temporal information.

\begin{figure}[ht]
\centering
\includegraphics[width=0.6\linewidth]{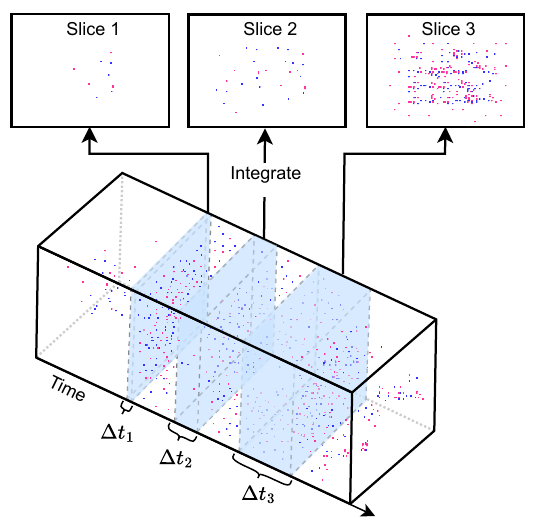}
\caption{Spatio-temporal data collected by DVS with variable temporal integration lengths.}
\label{fig:dvs integration}
\end{figure}

We process the N-MNIST dataset with varying temporal integration lengths to generate different frame-like sequence datasets. As shown in Figure~\ref{fig:dvs integration}, this approach generates multiple datasets, each of which is characterized by a specific temporal resolution determined by the temporal integration length, thereby enabling a comprehensive assessment of the generalization capability of SNNs. In our experiments, the N-MNIST event streams are integrated over different temporal integration lengths (i.e., 1$ms$, 2$ms$, 3$ms$, 5$ms$, and 10$ms$) to create varying temporal resolutions. The primary training is conducted on the 3$ms$ configuration, and the network structures and hyper-parameter settings are consistent with those mentioned in Table~\ref{tab:network structures} and Table~\ref{tab:hyperparameter} for the N-MNIST dataset. The corresponding RNN network structure is ``Input(2312)--RNN(512)--FC(10)--Output'', including Vanilla RNNs, Long Short-Term Memory (LSTM) \cite{hochreiter1997long} networks, and Gated Recurrent Units (GRU) \cite{cho2014learning}. After pre-training, subsequent testing is performed on datasets with different temporal integration lengths (i.e., 1$ms$, 2$ms$, 5$ms$, and 10$ms$) to evaluate the model's generalization capability by examining the testing accuracy.

\subsubsection{Results.}

\begin{table*}[ht]
\centering
\caption{Generalization comparison between models under variable timescales.}
\resizebox{\textwidth}{!}{%
\begin{tabular}{lccccccccc}
\br
& \multicolumn{3}{c}{Components} && \multicolumn{5}{c}{Temporal Resolution} \\
\cline{2-4} \cline{6-10}
Model & Leakage & Reset & Recurrence && 1ms & 2ms & 3ms & 5ms & 10ms \\
\mr
SNN (Baseline) & Normal & Normal & Without && 90.82\% & 94.14\% & 94.76\% & 94.80\% & 94.63\% \\
SNN & Without & Normal & Without && 88.70\% & 94.49\% & 95.18\% & 95.26\% & 95.07\% \\
SNN & Complete & Normal & Without && 90.98\% & 94.57\% & 95.03\% & 95.12\% & 95.14\% \\
SNN & Normal & Without & Without && 93.90\% & 95.83\% & 96.18\% & 96.30\% & 96.12\% \\
SNN & Normal & Normal & With && 75.51\% & 93.72\% & 96.21\% & 94.00\% & 80.41\% \\
\mr
SNN & Without & Normal & With && 59.14\% & 93.48\% & 96.48\% & 92.57\% & 70.11\% \\
SNN & Without & Without & With && 36.92\% & 89.16\% & 93.40\% & 90.91\% & 74.01\% \\
\mr
LSTM & - & - & - && 31.62\% & 90.52\% & 97.84\% & 63.10\% & 9.84\% \\
GRU & - & - & - && 23.25\% & 77.69\% & 96.88\% & 81.83\% & 45.20\% \\
Vanilla RNN & - & - & - && 17.68\% & 18.28\% & 93.84\% & 37.58\% & 19.92\% \\
\br
\end{tabular}}
\label{tab:model_accuracy}
\end{table*}

As reflected in Table~\ref{tab:model_accuracy}, an intuitive observation indicates a significant superiority in generalization performance for SNNs compared to all RNN models. In the spectrum of variant SNNs, it is noted that the absence of leakage leads to a decline in generalization, particularly evident under the 1msms temporal resolution. Furthermore, the incorporation of recurrence seems to impair the generalization capability notably. When combining the worst setting without leakage, without reset, and with recurrence in an SNN model, the generalization capability significantly drops, approaching the level observed in the LSTM model. It seems that the impact of the reset is more subtle compared to those of the leakage and recurrence, but helpful when the model accuracy is low for example comparing the settings without leakage, normal/without reset, and with recurrence. 

\subsubsection{Comparing Loss Landscapes.}

The concept of loss landscape flatness is an important concept for understanding the generalization capability of a neural network. Flat regions in the loss landscape indicate where small variations of the network parameters result in minor changes of the loss value \cite{bengio2012practical}. In contrast, steep regions represent sensitive areas where minor parameter changes can lead to significant loss alterations. Therefore, a flatter minima in the loss landscape implies that noises and shifts in the data distribution will not lead to significant loss increases, ensuring more stable performance for unseen data.

We adopt the method introduced by Li et al. \cite{li2018visualizing} to visualize the loss landscape of neural networks. This approach allows to visualize the high-dimensional loss landscape by projecting it onto a two-dimensional space. The key concept is to plot the network’s loss with respect to random directions in the parameter space. This visualization technique provides an intuitive understanding of the optimization landscape, highlighting areas with a flat minima for better generalization.

\begin{figure}[ht]
\centering
\includegraphics[width=\linewidth]{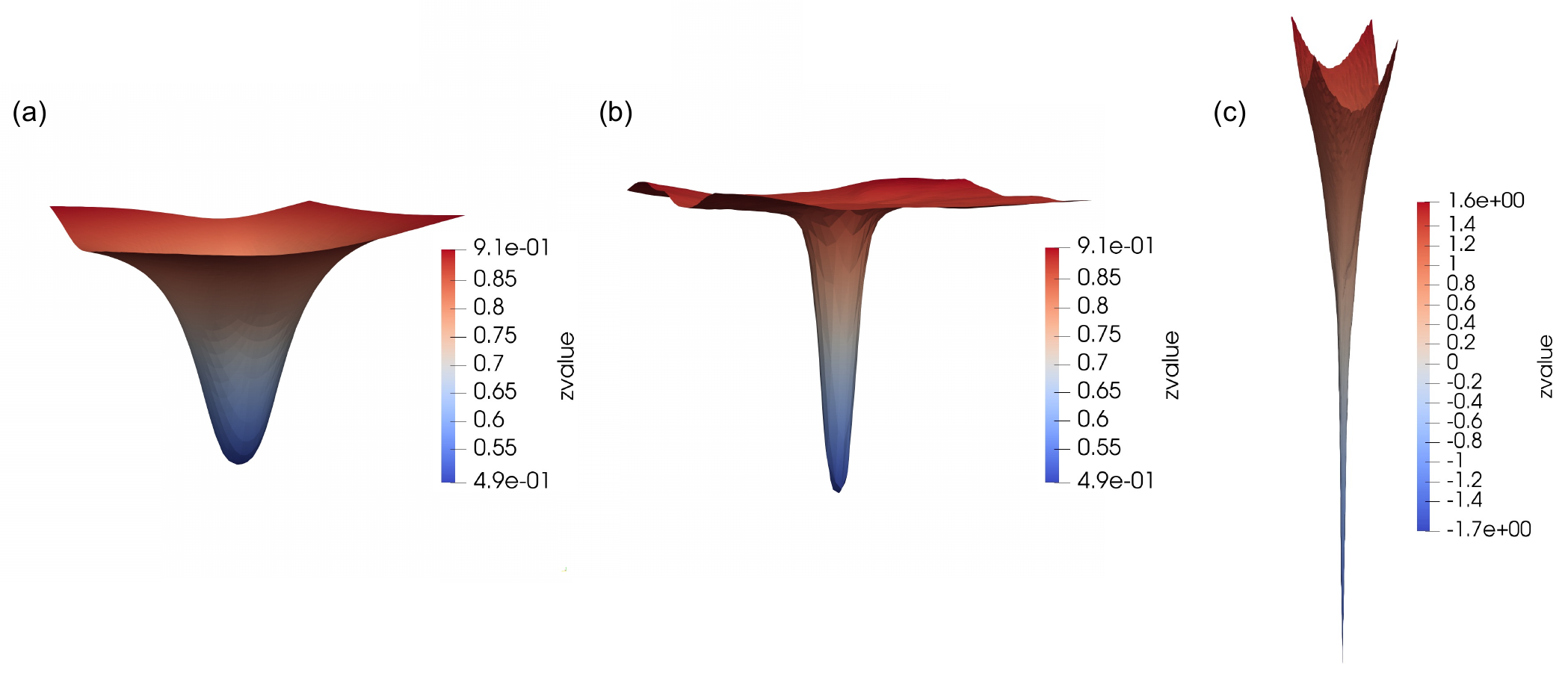}
\caption{Comparison of the loss landscapes: (a) vanilla SNN; (b) variant SNN without leakage and with recurrence; (c) LSTM.}
\label{fig:loss landscape}
\end{figure}

We visualize three models: a vanilla SNN, a variant SNN without leakage and with recurrence, and an LSTM model, whose network structures and hyper-parameter settings are consistent with those described in Table~\ref{tab:network structures} and Table~\ref{tab:hyperparameter} for the N-MNIST dataset. As presented in Figure~\ref{fig:loss landscape}, the loss landscape visualization reveals that the vanilla SNN exhibits the flattest optimization landscape, followed by the variant SNN, while the LSTM shows the steepest landscape. This suggests that SNNs inherently possess better generalization. It might be attributed to their spike-based neuronal dynamics with a discrete state space and automatic membrane potential decaying that can resist perturbations to a great extent. It can be seen that removing the leakage that prevents the membrane potential from decaying and introducing recurrence that increases the probability of state transfer in the state space, would harm the generalization capability. The flat loss landscape observed in SNNs implies that they have a higher tolerance to parameter or input perturbations, which can be advantageous in dynamic and noisy environments. With above observations, we further analyze the impacts of individual modelling components of SNNs on the generalization capability in the following parts.

\textbf{The role of the leakage.} In vanilla SNNs, leakage refers to the gradual decay of the membrane potential over time, preventing indefinite accumulation of the membrane potential error. Removing the leakage would magnify the impact of errors, especially in networks with long time steps or deep layers. In the context of BPTT, this can be illustrated by considering the gradient of the loss function with respect to the membrane potential at a given time step \( t \), denoted as \( \frac{\partial L}{\partial u(t)} \). In the variant without leakage, the gradient can be expanded as
\begin{equation}
\frac{\partial L}{\partial u(t)} = \frac{\partial L}{\partial o(t)} \cdot \frac{\partial o(t)}{\partial u(t)} + \frac{\partial L}{\partial u(t+1)} \cdot \frac{\partial u(t+1)}{\partial u(t)}.
\end{equation}
Note that \( \frac{\partial u(t+1)}{\partial u(t)} \) can be significantly impacted by the absence of leakage. Typically, a leakage coefficient \( k_\tau \) would scale down \( \frac{\partial u(t+1)}{\partial u(t)} \), reducing the propagated error. Without this scaling after removing the leakage, errors propagate fast, affecting the learning stability and generalization.

\textbf{The role of the reset mode.} The reset mechanism, which only occurs at the firing time step, has a lower impact on the error accumulation. Although the reset mechanism induces a discontinuity in the membrane potential, this effect does not inherently contribute a lot to the error propagation over multiple time steps. However, it's important to note that when combining the setting without reset to the setting without leakage and with recurrence, the generalization capability notably drops. This suggests that while the individual effect of the reset mode might be limited, its interaction with other components can lead to significant influences.

\textbf{The role of the recurrence pattern.} Recurrence in SNNs, introducing cross-neuron dynamics over time steps, allow these errors to propagate more broadly across the network. This exacerbates the error propagation, significantly impacting the generalization capability. This effect is especially pronounced when combining the absence of leakage together, where the errors from previous time steps are not attenuated. 

The recurrence mechanism can be formalized as
\begin{equation}
u(t) = f(\sum_j v_{i j}^n o_j^{t-1, n}, \ldots),
\end{equation}
where \( f \) represents the update function incorporating recurrent inputs. The gradient \( \frac{\partial L}{\partial u(t)} \) in this situation becomes more complex due to the additional terms involving past outputs of other neurons. This complexity magnifies the error accumulation, leading to more significant degradation of generalization.

\subsubsection{Feature Space Analysis with t-SNE Visualization.}

We further visualize the feature representation learned by each variant SNN using t-SNE dimensionality reduction \cite{van2008visualizing, wattenberg2016use}. This technique allows us to visualize the high-dimensional features from the last hidden layer in a two-dimensional space, providing an intuitive understanding of how different components influence the feature space.

\begin{figure}[ht]
\centering
\includegraphics[width=\linewidth]{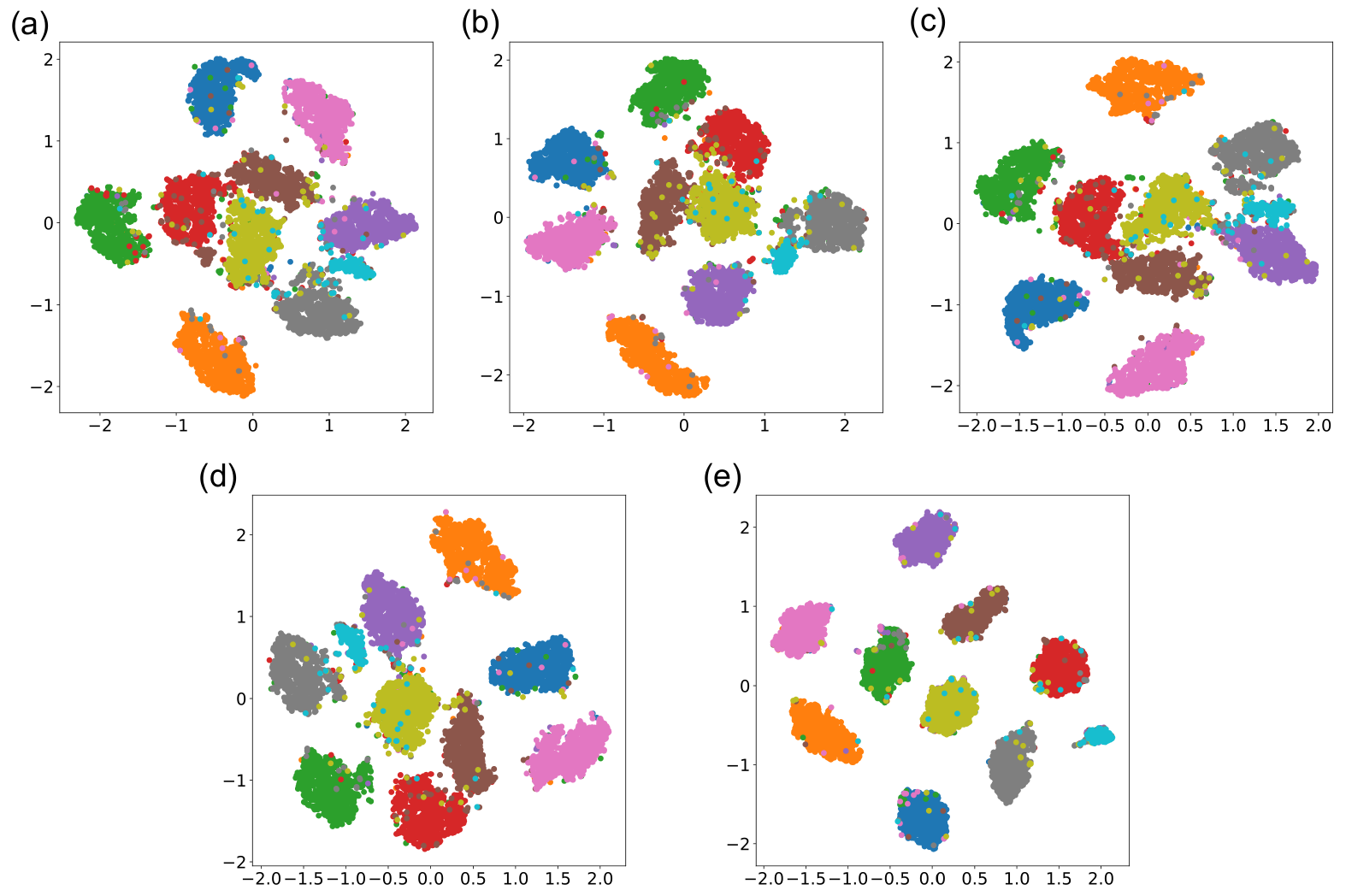}
\caption{Feature visualization of the last hidden layer of variant models: (a) vanilla SNN; (b) without leakage; (c) complete leakage; (d) without reset; (e) with recurrence.}
\label{fig:tsne}
\end{figure}

\begin{table}[ht]
\centering
\normalsize
\caption{Silhouette coefficients of variant models.}
\begin{tabular}{ll}
\hline
Variant & Silhouette Coefficient  \\
\hline
Baseline & 0.271 \\
Without leakage & 0.355 \\
Complete leakage & 0.273 \\
Without reset & 0.277 \\
With recurrence & 0.855 \\
\hline
\end{tabular}
\label{tab:Silhouette Coefficients}
\end{table}

As illustrated in Figure \ref{fig:tsne}, the variant SNN with recurrence demonstrates the most distinctive class separation, followed by closer class distributions observed in the remaining variants. We conducted a quantitative analysis using the silhouette coefficient \cite{rousseeuw1987silhouettes} with the cosine distance in the high-dimensional space, as shown in Table \ref{tab:Silhouette Coefficients}, which demonstrates the aforementioned perspective. This observation aligns with previous findings that the setting with recurrence can enrich the model capability in capturing complex temporal dynamics, resulting in more distinctive feature representation and higher accuracy. However, this leads to a steeper loss landscape with poorer generalization, which is usually ignored by most works focusing on the accuracy result.

\subsubsection{Comparison to RNNs.}

RNNs, with natural recurrence, face challenges in handling long sequences, primarily due to gradient vanishing or exploding. Unlike the vanilla RNN, LSTM and GRU typically include multiple state paths and gating mechanisms to mitigate this issue. Specifically, LSTM provides a more effective control over the information flow through two state variables and four gates, which allows to retain or forget information flexibly across long sequences. GRU further simplifies the architecture by using only two gates and merging the cell and hidden states. In a nutshell, their different architectures result in a performance priority following LSTMs, GRUs, and vanilla RNNs. SNNs remain the multiple state paths, membrane potential and spike event, which behave like LSTMs with great potential in learning long sequences. Furthermore, SNNs simplify the complex gate structures but introduce leakage, firing, and reset mechanisms that endow them enhanced generalization.

\subsection{Robustness against Adversarial Attack}

Adversarial attacks produce a significant challenge for neural networks, especially in applications where security and reliability are critical \cite{madry2017towards}. These attacks are characterized by malicious input perturbations that mislead neural networks to incorrect classifications or predictions. In this context, SNNs have attracted attention for robust intelligence due to their unique temporal dynamics and modelling components, which are distinct from conventional ANNs. Investigating the robustness of SNNs is not only academically fascinating but also holds substantial practical value in developing more secure intelligent systems.

\begin{figure}[ht]
\centering
\includegraphics[width=\linewidth]{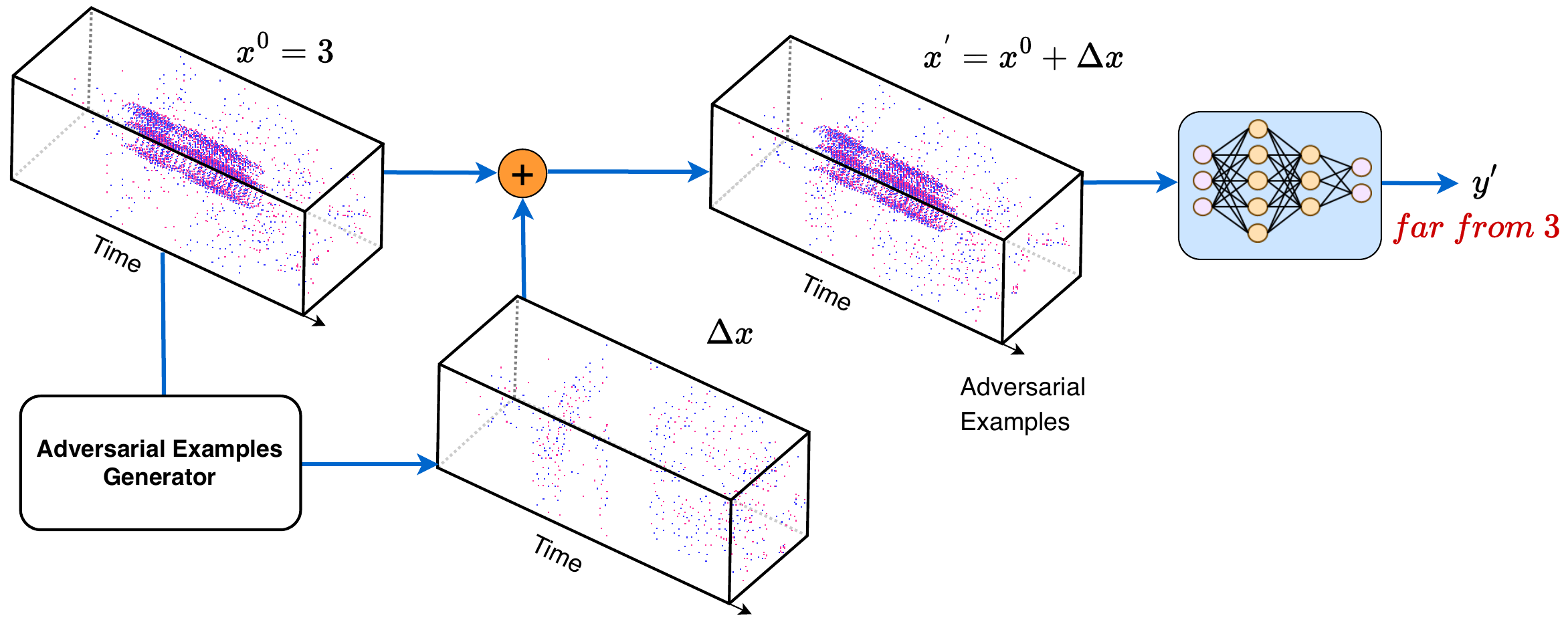}
\caption{Workflow of the adversarial attack for SNNs.}
\label{fig:adversarial attack}
\end{figure}

\subsubsection{Experimental Setup.}

Inspired by the approach in Liang et al. \cite{liang2021exploring}, we generate adversarial examples for SNNs following the workflow in Figure~\ref{fig:adversarial attack}. The attack involves perturbing the input spikes in a way that causes misclassification while ensuring that the perturbations are minimal and indiscernible to human observers. Network structures and hyper-parameter settings are consistent with those described in Table~\ref{tab:network structures} and Table~\ref{tab:hyperparameter} for the N-MNIST dataset. The primary metric for evaluating adversarial robustness is the attack success rate, which measures the proportion of adversarial examples that can lead to misclassification successfully. Furthermore, the model robustness can be quantified by the average perturbation magnitude required to induce misclassification. To establish a more fair comparison of robustness, we fix the attack success rate at 100\% and then compare the average perturbation needed to reach such high attack success rate. A smaller perturbation implies lower robustness. 

For drawing more general conclusions, we conduct extra adversarial attack experiments using the Eventdrop \cite{gu2021eventdrop, ma2023exploiting} method. This method specifically involves randomly dropping events from the N-MNIST dataset to create adversarial examples. The dropping probability is set by a parameter $\rho$. The dropping strategy we select is the random drop, which combines spatial and temporal event-dropping strategies.

\begin{table}[ht]
\centering
\normalsize
\caption{Accuracy under EventDrop attack.}
\begin{tabular}{ll}
\hline
Variant & Accuracy  \\
\hline
Baseline & 85.67\% \\
Without leakage & 77.51\% \\
Complete leakage & 86.66\% \\
Without reset & 88.83\% \\
With recurrence & 82.81\% \\
\hline
\end{tabular}
\label{tab:EventDrop}
\end{table}

\subsubsection{Results and Analysis.}

\begin{figure}[ht]
\centering
\includegraphics[width=\linewidth]{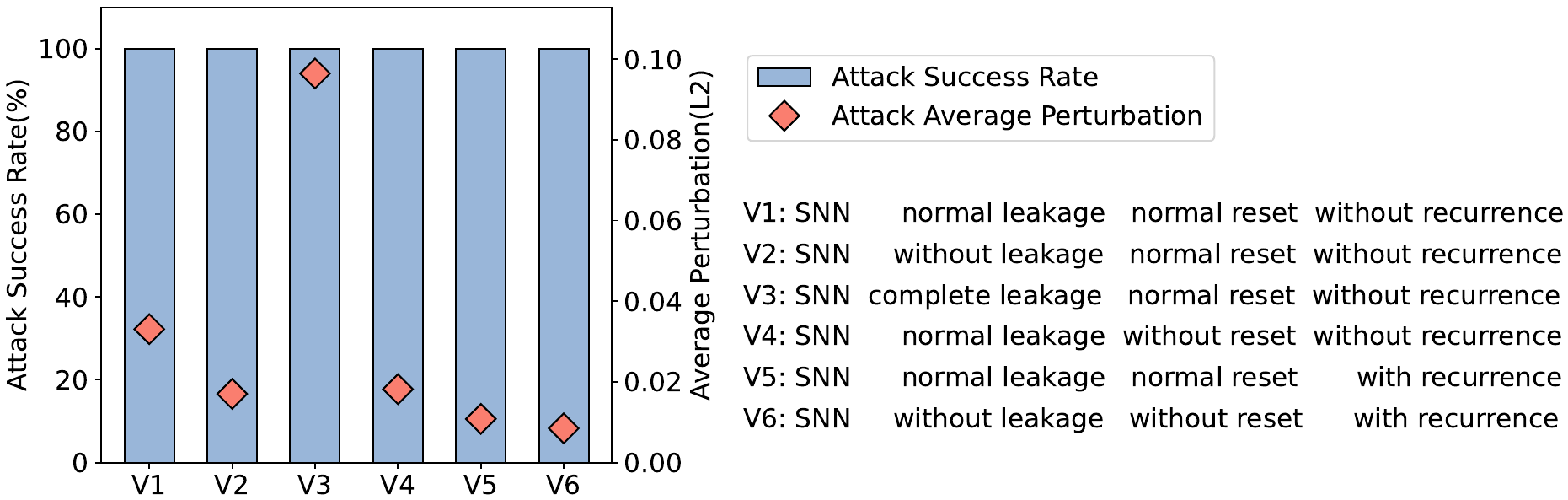}
\caption{Comparing adversarial robustness between variant models.}
\label{fig:robust}
\end{figure}

The results obtained from the adversarial robustness experiments are presented in Figure \ref{fig:robust} and Table \ref{tab:EventDrop}. Among the variant models, leakage and recurrence are the most influential factors that impact the adversarial robustness. As the leakage rate increases (i.e., smaller $k_{\tau}$), the robustness of SNNs against adversarial attack can be improved; conversely, SNNs with reduced or without leakage exhibit a decline in robustness. This phenomenon can be attributed to the fact that a higher leakage rate prevents the error accumulation, thereby reducing the network's susceptibility to small input perturbations. In addition, the introduction of recurrence tends to degrade the robustness. While the recurrent topology enhances the model capacity in processing temporal data, it also introduces additional pathways for the error propagation, making recurrent SNNs more vulnerable to adversarial attack. The influence of the reset component on adversarial robustness is less significant with slight variations under different attack methods, suggesting that the reset mechanism, which primarily affects the network at the moment of spike generation, does not heavily impact the overall resistance to adversarial attack.

Generalization experiments focus on testing the model performance on processing unseen data, while adversarial attack examines specific robustness of the model in resisting adversarial examples. Notably, the findings in adversarial attack experiments are consistent with those observed from the generalization experiments. This alignment further strengthens our understanding of the functional roles of SNN modelling components, paving the way for the development of more effective and robust neuromorphic models.

\section{Optimization Suggestions}

Based on the comprehensive experiments and analyses presented in previous sections, here we summarize several suggestions for optimizing SNNs in different tasks. These suggestions leverage our in-depth understanding of the functional roles of modelling components in the LIF-based SNNs.

\textbf{Suggestions for temporal computing tasks.} (1) For tasks that need a long-term memory and sensitive response, a properly set or even learnable leakage rate is critical. A too high or low leakage rate cannot handle temporal information well, and a learnable leakage rate is recommended. (2) For tasks that need continuous processing of temporal information without disruption such as the delayed spiking XOR problem, disabling the reset mechanism could be beneficial. However, the increase of the firing rate without reset would decrease the computational efficiency.  (3) For tasks that model complex temporal dynamics, the incorporation of recurrence can enhance the representation ability. However, this might lead to overfitting, for which considering the trade-off between accuracy and generalization/robustness is necessary.

\textbf{Suggestions for generalization and robustness.} (1) For tasks where generalization and robustness are paramount, a higher leakage rate (i.e., smaller $k_{\tau}$) can enhance the resistance to input perturbations by reducing error accumulation. (2) For these tasks, avoiding recurrence, although beneficial for temporal processing, is helpful for improving model generalization and robustness due to the increasing error propagation paths. Note that the gain of higher generalization and robustness might harm the application accuracy, which again reflects the trade-off mentioned above.


\section{Conclusion}

This work systematically explores the functional roles of modelling components in LIF-based SNNs. With customized variant models and extensive comprehensive on diverse benchmarks, we get valuable observations on how the leakage, reset, and recurrence components influence the behaviors of SNNs. Finally, we provide suggestions for model optimization in different tasks. Specifically, the leakage component plays a crucial role in application accuracy and model robustness. A appropriate leakage rate can enhance the capability in processing temporal information with a good balance between the long-term memory and sensitive response, and a low leakage usually makes the model sensible to noisy input perturbations due to the larger error accumulation. The reset component, while not impacting generalization and robustness significantly, sometimes degrades the application accuracy in tasks that need uninterrupted temporal processing and can be improved by disabling it for maintaining temporal continuity. The recurrence component allows to model complex temporal dynamics by introducing feedback connections, which can improve the application accuracy for complex temporal computing tasks. However, the recurrence increases the risks of worse generalization and robustness due to the cross-neuron error propagation paths. There findings deepen the understanding of SNNs and help identify the key modelling components for guiding the development of effective and robust neuromorphic models in different application scenarios.

\section*{Data availability}
All data used in this paper are publicly available and can be accessed at \url{https://zenkelab.org/resources/spiking-heidelberg-datasets-shd/} for SHD and SSC datasets, \url{https://www.garrickorchard.com/datasets/n-mnist} for N-MNIST dataset, \url{https://ibm.ent.box.com/s/3hiq58ww1pbbjrinh367ykfdf60xsfm8/folder/50167556794} for DVS Gesture dataset, \url{http://yann.lecun.com/exdb/mnist/} for MNIST dataset, \url{https://www.cs.toronto.edu/~kriz/cifar.html} for CIFAR-10 dataset.

\section*{Acknowledgments}

This work was partially supported by National Natural Science Foundation of China (No. 62276151, 62106119), Key-Area Research and Development Program of Guangdong Province (No. 2021B0909060002), CETC Haikang Group-Brain Inspired Computing Joint Research Center, and Chinese Institute for Brain Research, Beijing.

\section*{References}
\bibliography{ref}

\end{document}